\documentclass{article}

\usepackage[final,nonatbib]{neurips_2022}

\usepackage[utf8]{inputenc} 
\usepackage[T1]{fontenc}    
\usepackage{hyperref}       
\usepackage{url}            
\usepackage{booktabs}       
\usepackage{amsfonts}       
\usepackage{nicefrac}       
\usepackage{microtype}      
\usepackage{xcolor}         

\usepackage{graphicx}

\title{Privacy-Preserving Machine Learning for Collaborative Data Sharing via Auto-encoder Latent Space Embeddings}
\author{%
  Ana María Quintero Ossa \\
  Industrial Engineering, Los Andes University\\
  Bogotá, Colombia \\
  \texttt{am.quintero12@uniandes.edu.co} \\
  \AND
  Jesús Solano \\
  Artificial Intelligence, Rappi\\
  Bogotá, Colombia \\
  \texttt{jesus.solano@rappi.com} \\
  \And
  Hernán García \\
  Artificial Intelligence, Rappi\\
  Bogotá, Colombia \\
  \texttt{javier.garcia@rappi.com} \\
  \And
  David Zarruk \\
  Statistics \& Machine Learning, Amazon\\
  Bogotá, Colombia \\
  \texttt{davidzarruk@gmail.com} \\
  \AND
  Alejandro Correa-Bahnsen \\
  Artificial Intelligence, Rappi\\
  Bogotá, Colombia \\
  \texttt{alejandro.correa@rappi.com} \\
  \AND
  Carlos Valencia \\
 Industrial Engineering, Los Andes University\\
  Bogotá, Colombia \\
  \texttt{cf.valencia@uniandes.edu.co} \\
  }

\begin{document}

\maketitle

\begin{abstract}
Privacy-preserving machine learning in data-sharing processes is an ever-critical task that enables collaborative training of Machine Learning (ML) models without the need to share the original data sources. It is especially relevant when an organization must assure that sensitive data remains private throughout the whole ML pipeline, i.e., training and inference phases. This paper presents an innovative framework that uses Representation Learning via autoencoders to generate privacy-preserving embedded data. Thus, organizations can share the data representation to increase machine learning models' performance in scenarios with more than one data source for a shared predictive downstream task. 
\end{abstract}

\section{Introduction}
Artificial Intelligence collaboration, as data sharing strategies, are standard practices that organizations are starting to use with each other to improve prediction model performance, increase the reliability of data, and, thus, acquire competitive data-based advantages \cite{sim2020collaborative} . Yet, in some use cases of real life, the data sharing process might not be possible because of privacy policies or intellectual property laws, even if the communication infrastructure between peers is safe \cite{kop2020machine}. Imagine for instance, two companies where each of them has a particular set of variables for the same group of users. In this scenario, both peers could take advantage of the complementary information available in the other company to predict a response variable (model output) and use it in decision-making processes. However,  given that there is sensitive data from users on both ends, the data sharing process is canceled, and so is the possibility of increasing the model performance. In this case, developing a strategy that allows them to share their information without losing the predictive power over the downstream variable would be appropriate for both organizations' ML model development. 

In that regard, academia and private organizations have developed multiple solutions and frameworks enabling data sharing based on technology and machine learning approaches. Most of these approaches are based on cryptography (e.g., homomorphic encryption \cite{Moore2014,aono2017privacy}), raw data perturbation (e.g., differential privacy \cite{Abadi2016}, local differential privacy \cite{Wang2021}, dimensionality reduction\cite{Chen2021}), and distributed architecture (e.g., federated learning) \cite{Al-Rubaie2018,nguyen2020autogan}. Notice that these solutions only preserve privacy in communication, change individual observation patterns, and present high maintenance requirements. Having said that, we are interested in building a privacy-preserving framework using recent deep learning modeling that will allow collaborative peers to share their data without losing the predictive power of the original features. 

This paper presents an innovative framework using representation learning via auto-encoders to create privacy-preserving embeddings of sensitive information and allow multiple data sources to collaborate in trustful machine learning model development. Additionally, we applied the proposed framework to three scenarios to conclude the applicability. The work can be summarized as follows: First, we overview available case studies on privacy-preserving machine learning to understand their limitations and the room for improvement. Then we go deep into the proposed method and stages of the general process to test the methodology. Subsequently, we introduce and develop the selected case studies and present their results. Finally, we conclude and present insights for future work.

\section{Background}
Since our proposed method proposes a novel methodology for privacy-preserving machine learning for collaborative model development using deep-learning autoencoders, we will understand traditional privacy-preserving approaches and representation learning to conclude about their possible integration as a solution to the problem we are addressing

\subsection{Privacy-preserving Machine Learning}
Privacy-preserving machine learning is an application field in the AI ecosystem which intends to close the gap between data ownership rights and the benefits of applying machine learning models with this data. Specifically, these models can protect the data or the developed model \cite{xu2021privacy}. Since we propose a strategy to allow peers to share data safely, we will go deep into data-oriented privacy guarantee applications. 

There are three main traditional approaches to solve the problem of privacy preserving ML. First, Encryption-Based Privacy-Preserving, which prevents data leakage between peers by transforming the feature set into a ciphertext, that can be analyzed as the original data \cite{10.1145/3214303}. Despite the security benefits of using these frameworks, such as Homomorphic Encryption, these solutions are limited when implementing the methodology in real-life scenarios because of technology requirements. On the other hand, we have architecture-based approaches, such as Federated Learning, which create a decentralized model development pipeline with data residing in multiple peers as mobile devices \cite{MLSYS2019_bd686fd6}. This solution is a practical approach when many contributors share the same information, but cannot be used when different peers share different information, which does not solve the problem we are addressing.

The third traditional approach to data-oriented privacy preservation is a perturbation of the original features. In particular, differential privacy is a commonly used strategy that takes advantage of the data distribution to mask individual observation values \cite{friedman2010data}; however, it might add significant noise to the original data, decreasing the data utility. Finally, applying dimensionality reduction to the original data can preserve the variance of each observation while obfuscating the original features. One way to apply this strategy is Principal Component Analysis, which creates a representation vector of the data. These new features can be used on the downstream model of interest. However, linear transformation for dimensionality reduction might lose other data relationships. Nguyen \textit{et.al.} \cite{nguyen2020autogan}, in their work AutoGAN-based Dimension Reduction for Privacy Preservation, used representation learning to preserve the privacy of images and include their embedding on anomaly detection.

\subsection{Representation Learning}
On the other hand, Representation Learning is a study field of Deep Learning that allows algorithms to learn representations of input data automatically. These techniques are widely used in alternative data such as images, speech, or text, and their applications include anomaly detection, pattern recognition, and dimensionality reduction. Autoencoders are neural networks specifically trained to encode input data and use this embedding to reconstruct the original data set with a minimum error\cite{10.5555/3045796.3045801}]. These neural networks are built with two structures: an encoder and a decoder. They are connected through the latent space representation of the data, which is the embedded vector of the original data \cite{sewak2020overview}. Additionally, representation Learning is widely used as a principal dimensionality reduction strategy since it structures a supervised machine learning model that seeks to find the best nonlinear features combination that represents the original data \cite{huang2020deep}. In this way, the latent space representation will be an abstract multi-dimensional space that encodes the original feature set but holds the proximity between observations that look alike.

Notice that privacy-preserving and representation learning are complementary research areas since the second one offers a deep learning strategy to encode data while keeping its core information and observation representation. In this way, this combination allows us to solve our main objective: achieve trustful data sharing between collaborative peers for machine learning model development. 

\section{Privacy-Preserving Machine Learning for Collaborative Data Sharing via Auto-encoder Latent Space Embeddings} \label{secc:method}

\begin{figure*}[!ht]
  \centering
  \includegraphics[width=100mm]{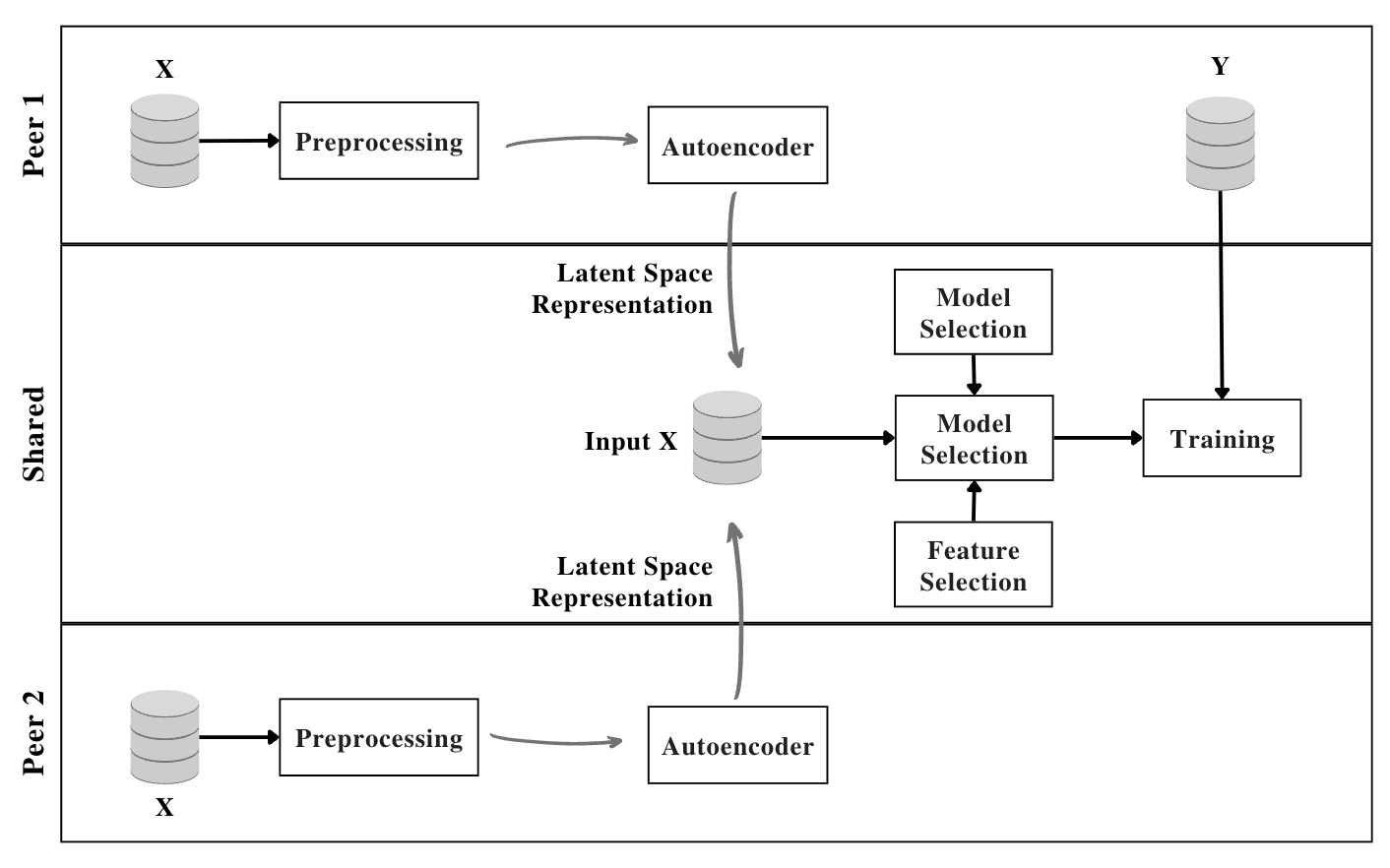}
  \caption{Privacy-Preserving Machine Learning Framework: the two peers share the respective latent representation of the corresponding database to a common data source that is then used to train the Machine Learning model in the \textit{training phase}}. 
  \label{fig:0}
\end{figure*}

We propose a method that seeks multiple peers could use representation learning to embed their data as a privacy-preserving strategy and share it among them without losing predictive power. Proving that this strategy works to share data trustfully and keeps the predictive model's performance will add another possibility for organizations to have AI collaboration practices. Our framework displayed in Fig. \ref{fig:0},  considers multiple data sources willing to contribute to each other by sharing the data but must respect the privacy of sensitive data. Notice that they collaborate on complementing the feature set of an observation identified by a standard ID.


In traditional data-sharing pipelines to train collaborative machine learning models, both ends contribute by sharing a raw dataset that will be merged with a standard ID in all observations. After the appropriate data preprocessing, one peer (or both of them) can train a machine-learning model with a more extensive feature set that can improve the predictive power over the objective variable. Unlike this approach, we propose to include an additional step previous to the data merging, in which peers will obtain a latent space representation of the original data, in other words: getting an obfuscated dataset ready to be shared. In this way, peers will join each data representation to train a shared supervised downstream task to predict the same objective variable without losing predictive power and intent to improve the overall performance by sharing their data.

For the initial scope of this method, we limit the number of peers to two, but this strategy could scale to more than this quantity. Additionally, we assume that the involved peers will share a representation of the whole feature set. However, for applications in real-life scenarios, there may be no need for both to apply privacy-preserving strategies.


\section{Evaluation}
\subsection{Data Sets}

To test the proposed framework as well as define scenarios likely to occur in real-life use cases, we selected three public data sources: House Pricing~\cite{ahmed2016house}, Mnist Numbers~\cite{deng2012mnist}, and Buzz in Social Media ~\cite{buzz}. We choose these data sets to test the performance of our framework under different characteristics and ensure we include possible scenarios, thus, being able to have a generalized framework. This way, we guarantee that we have both prediction tasks, regression, and classification. Additionally, we considered the variations in the available features by dimension and type. 
In other words, we seek to include different downstream tasks, dimensionality, and feature types to test the robustness and scalability of the solution. 

\begin{table} [!ht]
  \label{sample-table}
  \centering
  \begin{tabular}{llll}
    \toprule

    Data set     & Num. Observations     & Num. Features & Prediction Downstream Task \\
    \midrule
    House Pricing ~\cite{ahmed2016house}  & 21613 & 12  &  Regression \\
    Mnist Numbers ~\cite{deng2012mnist}     & 35000 &  784  &  Multi Class Classification\\
    Buzz in Social Media ~\cite{buzz}  & 87488  & 77 & Regression \\
    \bottomrule
  \end{tabular}
\caption{This table shows the widely-known benchmark data sets that we used to test our privacy-preserving framework. The number of total samples, number of features, and Machine Learning tasks performed over each dataset validates are correspondingly reported.}
\end{table}

\subsection{Experiments}
We set up a baseline model without the privacy-preserving strategy and four privacy-preserving scenarios to guarantee reliable and comparable results based on the proposed method in Section \ref{secc:method}.

\textbf{Scenario 0 | Baseline}\label{scenario:s0} Trains a predictive model for the downstream task using a single data source, which is considered as the raw dataset in this work. In this scenario, we train a traditional supervised machine learning model and include randomized search as a hyperparameter tunning strategy. The performance of this baseline model will be the performance that we seek to maintain in the following scenarios.

\textbf{Scenario 1 | Representation Learning with a single shared autoencoder} \label{scenario:s1} In this case, we preprocess a unique dataset to obtain a single representation vector and use it to train a predictive model for the downstream task. Thus, evaluate the predictive performance of an accurate representation.

\textbf{Scenario 2 | Representation Learning with individual autoencoders}\label{scenario:s2} Simulate two peers by splitting the initial data set and preprocessing them individually to obtain a representation vector for each source. To train the predictive model for the downstream task, we join those vectors using the observations' ids.

\begin{figure*}[!ht]
  \centering
  \includegraphics[width=100mm]{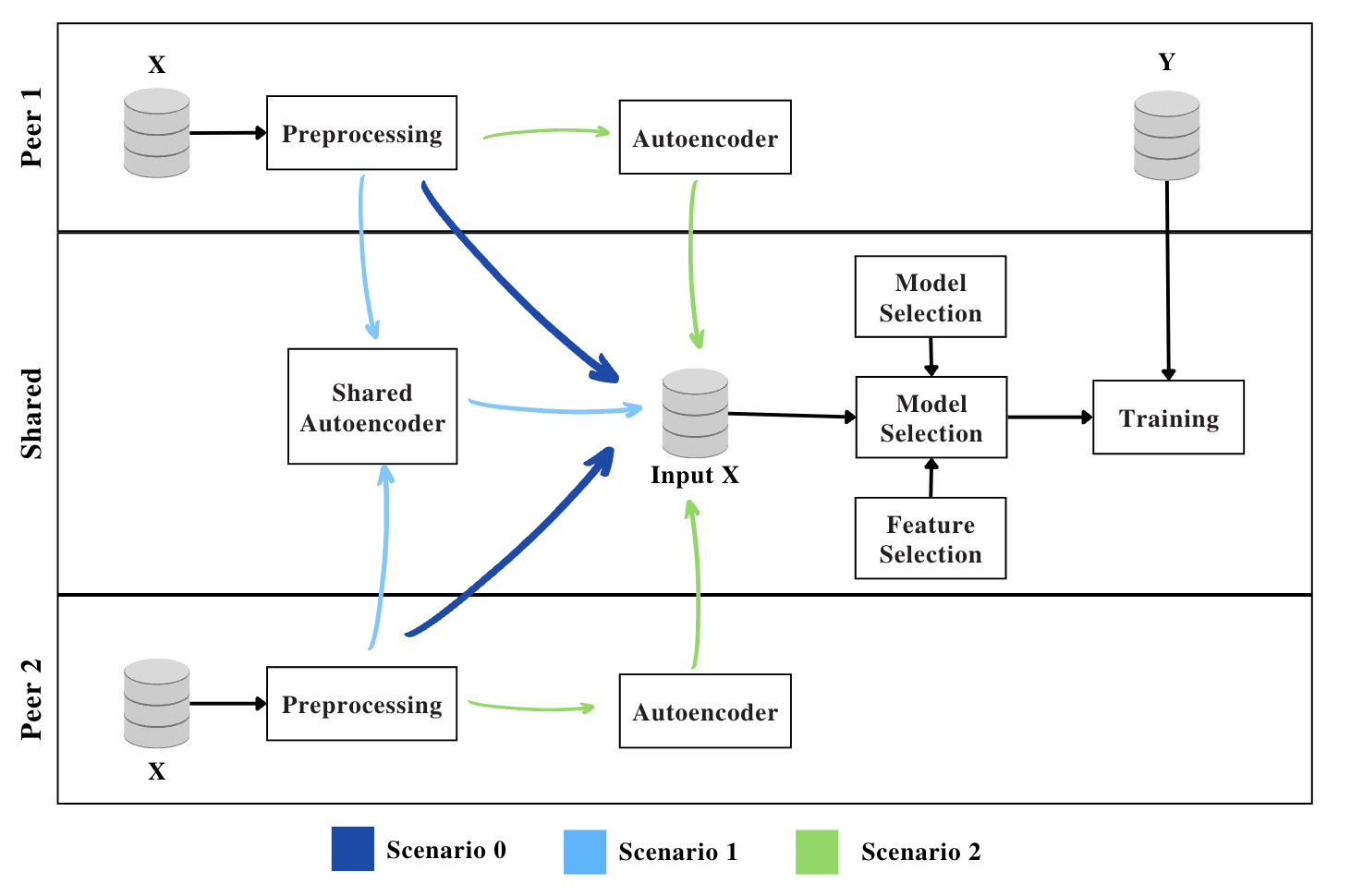}
  \caption{General Autoencoder structure for the \textit{training phase} and respective scenarios explored in this work, namely (1) Scenario 0: Baseline, Representation Learning with (2) Scenario 1: single shared autoencoder, and (3) Scenario 2: individual autoencoders per each peer.}
  \label{fig:2}
\end{figure*}

The following diagram presents the pipeline of the data preprocess and sharing between both peers \ref{fig:2}

We design the following two scenarios taking into account the hypothesis that the autoencoder could follow a non-naive approach when estimating the latent space representation and improve the downstream task metrics.

\textbf{Scenario 3 | Representation Learning with shared autoencoder non-naive approach}\label{scenario:s3}  We test the impact on the downstream task if the autoencoder model considers the predictive variable on the principal model. In this scenario, we adjust the autoencoder and transform it into a multitask neural network that, on one side, predicts the representation performance and, on the other, predicts the objective variable. Taking into account that both peers have access to the predictive variable, we replicate the second scenario but update the encoder stage of the process as mentioned above 

\textbf{Scenario 4 | Representation Learning with individual autoencoder non-naive approach}\label{scenario:s4}  We test the impact on the downstream task if the autoencoder model considers the predictive variable on the principal model. In this scenario, we adjust the autoencoder and transform it into a multitask neural network that, on one side, predicts the representation performance and, on the other, predicts the objective variable. Taking into account that both peers have access to the predictive variable, we replicate the second scenario but update the encoder stage of the process as mentioned above

\subsection{Experimental Setup}
\begin{figure*}[!ht]
  \centering
  \includegraphics[width=100mm]{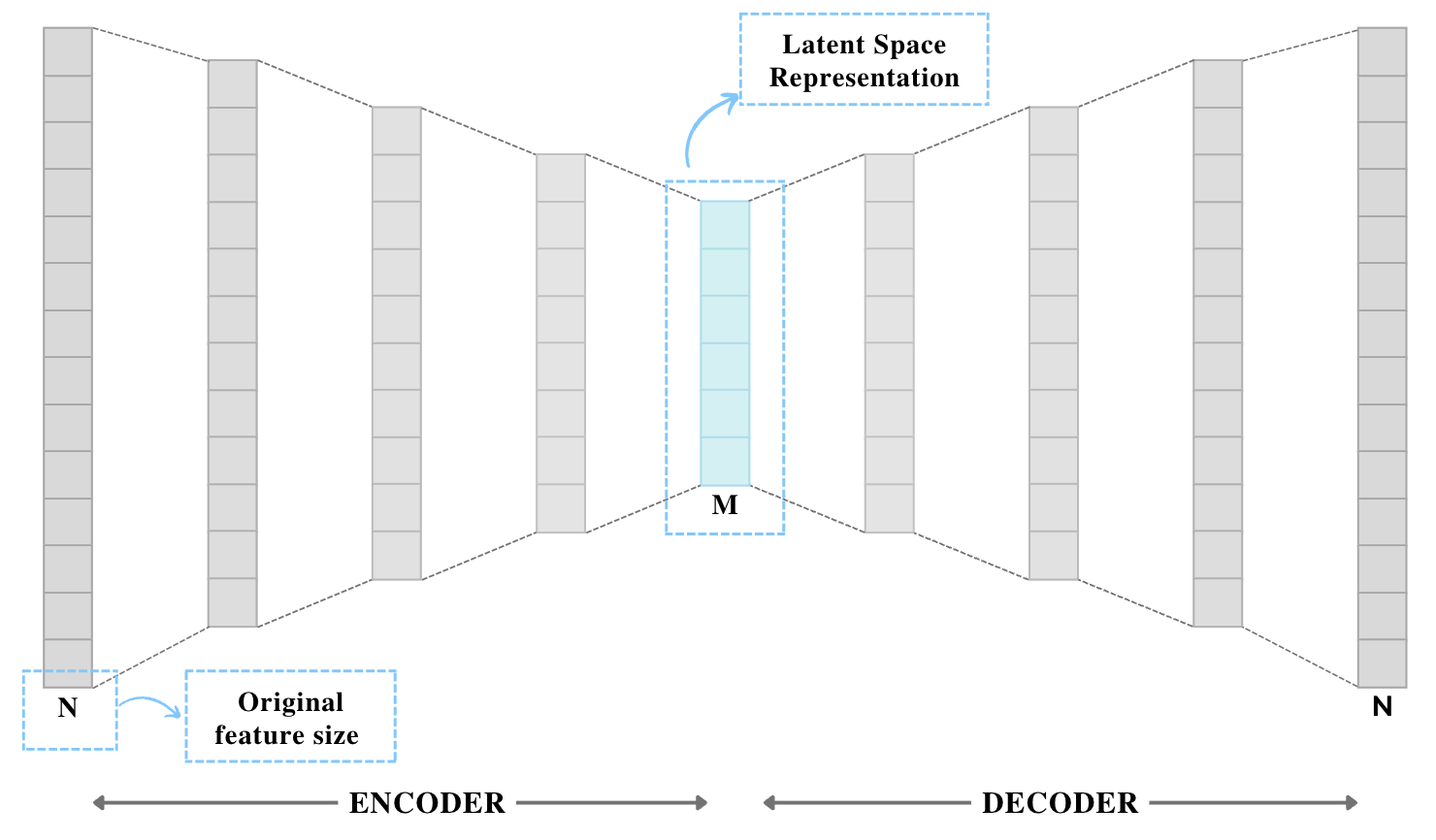}
  \caption{General Autoencoder Structure}
  \label{fig:3}
\end{figure*}

\textbf{Autoencoder Setup.} We use the Tensorflow framework to structure the Autoencoder Neural Network. Additionally, we defined one fixed structure for all experiments to be able to conclude over the general framework and not about the complexity of the network. In this case, the Autoencoder model presents two main components: encoder and decoder, each has four layers, and the connection between both is the latent space representation layer. Fig. \ref{fig:3} illustrates the general Autoencoder structure, where N represents the original feature size and M represents the embedding size.

Encoder: The encoder presents four layers. First is the input layer, which will have as many neurons as the original data frame features (N). Then the following three hidden layers will perform nonlinear transformation with dimensionality decrease between each other, reaching the final layer, which is the latent space representation and final layer. We select the dimensions N (for the input layer), [128, 64, 40] for the hidden layers, and M for the latent space representation size.

Decoder: Since the decoder is a mirror representation of the encoder, it will begin its structure from the latent space representation size (M). The following three layers will replicate the design of the encoder inversely to reach the final layer that will correspond to the original feature size (N). We select the dimensions M, [128, 64, 40] for the hidden layers and N for the final output size.

In additional considerations: 
Because of the type of input data that our framework use, we used ReLU as an activation function for every layer. Additionally, because of this same argument, and taking into account that we scale the input data, the loss function of the Autoencoder is  Mean Absolute Error. Finally, we selected an Adam Optimizer with a learning rate = 0.0001. 

\subsection{Technical Setup}
The following table presents the hardware specifications for training and testing the framework for all scenarios.

\begin{table} [!ht]
  \label{sample-table}
  \centering
  \begin{tabular}{ll}
    \toprule

    Parameter     & Technical Specs \\
    \midrule
    CPU Model  & Intel(R) Xeon(R) \\
    CPU Freq.    & 2.30 GHz\\
    No. CPU Cores  & 2\\
    GPU  & Nvidia K80 / T4\\
    GPU Memory & 12 GB\\
    No. CPU Cores  & 2\\
    Available RAM  & 12 GB\\
    \bottomrule
  \end{tabular}
\caption{This table lists the resources and their specifications used to test the proposed framework}
\end{table}

\section{Results}
\subsection{Encoding performance}

\textbf{Shared autoencoder}

We trained the autoencoder model with the complete datasets to test some previously mentioned scenarios. We tested the representation accuracy with the loss function of the autoencoder model and complementary metric of \textit{average correctly estimated observations per feature}, which is defined as an observation with less than 5\% MAPE. For House Pricing, the representation error is 5\%, and the average estimated observations per feature rate are 98\%. For Minst, the representation error is 7\%, and the average estimated observations per feature rate are 96\%. For Buzz in Social Media, the representation error is 6\%, and the average estimated observations per feature rate are 97\%.

\textbf{Individual autoencoders}

We trained one autoencoder model for each simulated data source to test some previously mentioned scenarios. As in the shared autoencoder, we tested the representation accuracy with the loss function of the autoencoder model and complementary metric of \textit{average correctly estimated observations per feature}. For House Pricing, the average representation error is 11\%, and the average estimated observations per feature rate are 86\%. For Mnist, the average representation error is 9\%, and the average estimated observations per feature rate are 94\%. For Buzz in Social Media, the representation error is 8\%, and the average estimated observations per feature rate are 94\%. 
\subsection{Representation Learning - Framework}

\textbf{House Pricing}

The downstream task for this experiment is to estimate the price in USD of a house, given some characteristics. We select an XGBoost Regressor model to predict the downstream task; additionally, we include hyperparameter tunning with Randomized Search Cross Validation, with this set of parameters: \textit{learning rate, max depth, min child weight, gamma, colsample bytree}.

\begin{table}[!ht]
\caption{House Pricing Scenarios Metrics}
  \label{sample-table}
  \centering
\begin{tabular}{c | c || c | c | c | c | c}
\toprule
  &   Metrics & 
  \begin{tabular}[c]{@{}c@{}}Scenario\\0\end{tabular} & \begin{tabular}[c]{@{}c@{}}Scenario\\1\end{tabular} & \begin{tabular}[c]{@{}c@{}}Scenario\\2\end{tabular}& \begin{tabular}[c]{@{}c@{}}Scenario\\3\end{tabular}& \begin{tabular}[c]{@{}c@{}}Scenario\\4\end{tabular}\\
\midrule \midrule
     &   R2 &  $90.26\%$  &  $84.26\% $ & $89.30\%$ &$89.79\%$ & $88.78\%$ \\
Train &   MAPE & $15.31\%$ &  $18.03\%$ & $16.03\% $ &$15.69\%$ & $17.74\%$\\
 \midrule                                                 
        &   R2 &   $90.32\%$ &  $84.41\%$ & $89.30\%$ &$88.93\%$ & $87.26\%$ \\
Validation  &   MAPE &  $14.88\%$ & $17.91\%$ &$15.89\%$ &$15.39\%$  &$16.89\%$ \\

  \midrule                                               
            &   R2 &  $90.29\%$ &  $84.27\%$ & 89.33\% & $89.21\%$ & $87.36\%$\\
Test        &   MAPE &  15.09\% &  17.97\% & 15.96\%  & $15.27\%$  &$17.58\%$ \\
\bottomrule
\end{tabular}
\end{table}

The results conclude that even with dimensionality augmentation (because of the limited number of features of this dataset), the latent space representation presents a low loss on the predictive power and that the downstream model can still predict the objective variable correctly. When the principal dataset simulates two data sources, the performance with both latent space representations reaches a high-level performance comparable to scenario one results. \hfill \break

\textbf{Mnist Numbers}

The downstream task for this experiment is to predict which number between 0 and 9 correspond to an image. We select a Multinomial Logistic Regression model to predict the downstream task. For this particular case, we transform the images so they can be used as tabular data

\begin{table}[!ht]
\caption{Mnist Scenarios Metrics}
  \label{sample-table}
  \centering
\begin{tabular}{c | c || c | c | c | c | c}
\toprule
  &   Metrics & 
  \begin{tabular}[c]{@{}c@{}}Scenario\\0\end{tabular} & \begin{tabular}[c]{@{}c@{}}Scenario\\1\end{tabular} & \begin{tabular}[c]{@{}c@{}}Scenario\\2\end{tabular}& \begin{tabular}[c]{@{}c@{}}Scenario\\3\end{tabular}& \begin{tabular}[c]{@{}c@{}}Scenario\\4\end{tabular}\\
\midrule \midrule
     &   Accuracy &  $94\%$  &  $88\%$ & $84\%$  & $92\%$  & $85\%$ \\
Train &   Precision & $94\%$ &  $88\%$ & $84\%$  & $92\%$ & $85\%$\\
     &   Recall &   $94\%$    &   $88\%$ & $84\%$  & $92\%$& $85\%$\\
 \midrule                                                 
        &   Accuracy &   $92\%$ &  $88\%$ & $84\%$  &  $91\%$ & $84\%$\\
Validation  &   Precision &  $92\%$ & $88\%$ &  $84\%$  & $91\%$& $84\%$\\
        &   Recall &  $92\%$ & $88\%$  & $84\%$  & $91\%$&$84\%$\\

  \midrule                                               
            &   Accuracy &  $92\%$ &  $88\%$ & $84\%$  & $91\%$ & $84\%$\\
Test        &   Precision &  $92\%$&  $88\%$ & $84\%$   & $91\%$ & $84\%$ \\
            &   Recall &   $92\%$    &  $88\%$ & $84\%$ & $91\%$ & $84\%$ \\
\bottomrule
\end{tabular}
\end{table}

The results conclude that even with dimensionality reduction, the latent space representation presents a low loss on the predictive power and that the downstream model can still predict the objective variable correctly. These results were expected taking into account the data's quality, their size and sparsity allows stable behavior. \hfill \break

\textbf{Buzz in social media}

The downstream task for this experiment is to predict the buzz for a tweet given some characteristics.  We select an XGBoost Regressor model to predict the downstream task; additionally, we include hyperparameter tunning with Randomized Search Cross Validation, with this set of parameters: \textit{learning rate, max depth, min child weight, gamma, colsample bytree}.

\begin{table}[!ht]
\caption{Buzz in social media Scenarios Metrics}
  \label{sample-table}
  \centering
\begin{tabular}{c | c || c | c | c | c | c}
\toprule
  &   Metrics & 
  \begin{tabular}[c]{@{}c@{}}Scenario\\0\end{tabular} & \begin{tabular}[c]{@{}c@{}}Scenario\\1\end{tabular} & \begin{tabular}[c]{@{}c@{}}Scenario\\2\end{tabular}& \begin{tabular}[c]{@{}c@{}}Scenario\\3\end{tabular}& \begin{tabular}[c]{@{}c@{}}Scenario\\4\end{tabular}\\
\midrule \midrule
     &   R2 &  $96.17\%$  &  $91.13\% $ & $89.28\%$ &$94.21\%$ & $89.01\%$\\
Train &   MAPE & $22.25\%$ &  $28.46\%$ & $32.18\% $ & $25.38\% $&$30.27\% $\\
 \midrule                                                 
        &   R2 &   $96.14\%$ &  $91.08\%$ & $89.68\%$ & $93.87\%$&$88.47\%$\\
Validation  &   MAPE &  $24.76\%$ & $28.51\%$  &  $33.42\%$  & $26.19\% $&$31.06\% $\\

  \midrule                                               
            &   R2 &  $96.19\%$ &  $91.55\%$ & 89.01\% &$94.03\%$ &$87.90\%$\\
Test        &   MAPE &  $23.87\%$ &  28.94\% & 33.23\% & $25.94\% $&$31.89\% $ \\
\bottomrule
\end{tabular}
\end{table}

The results conclude that, as in the Mnist case study, even with dimensionality reduction, the latent space representation presents a low loss on the predictive power and that the downstream model can still predict the objective variable correctly. Unlike the House Prices performance, this data frame presents a more significant difference from scenario 0, which is a consequence of the number of variables included.

\section{Conclusions \& Future Work}
In this paper, we propose an alternate solution to traditional privacy-preserving approaches in machine learning and proof that with an accurate representation learning model, peers can share an embedded dataset that follows the observations' patterns and behavior. Changing the original features to a latent space representation does not drastically deteriorate the performance of the downstream task. In our use cases, the model results decreased for less than 10pp with a representation error between 5\% and 11\%. Therefore, peers or organizations can collaborate without risking the organization's privacy policies or violating potential clients' privacy concerns. 

For future considerations, each data source should develop a custom autoencoder neural network implementation to improve the representation performance and guarantee that it fulfills the dataset requirements. In addition, even if we assume that dimensionality reduction keeps data privacy, we will develop measurements to quantify the privacy level for each dataset. This measure should take into account the complexity of the embedding and the difficulty for attackers to decode the original dataset. Finally, we will test this framework with organizational data from different sources and conclude over a real-life scenario.




\end{document}